\documentclass{article}
\usepackage[utf8]{inputenc}
\usepackage{authblk}
\usepackage{setspace}
\usepackage[margin=1.25in]{geometry}
\usepackage{graphicx}
\graphicspath{ {./figures/} }
\usepackage{subcaption}
\usepackage{amsmath}
\usepackage{booktabs}
\usepackage{wrapfig}
\usepackage{amssymb}
\usepackage{mathtools}
\usepackage{amsthm}
\usepackage{url}
\usepackage{xcolor}
\usepackage{hyperref}

\title{Transformer Conformal Prediction for Time Series}

\author[1]{Junghwan Lee}
\author[1]{Chen Xu}
\author[1]{Yao Xie\footnote{Correspondence: yao.xie@isye.gatech.edu}}

\affil[1]{H. Milton Stewart School of Industrial and Systems Engineering, Georgia Institute of Technology}

\date{}

\begin{document}

\maketitle

\begin{abstract}
We present a conformal prediction method for time series using the Transformer architecture to capture long-memory and long-range dependencies. Specifically, we use the Transformer decoder as a conditional quantile estimator to predict the quantiles of prediction residuals, which are used to estimate the prediction interval. We hypothesize that the Transformer decoder benefits the estimation of the prediction interval by learning temporal dependencies across past prediction residuals. Our comprehensive experiments using simulated and real data empirically demonstrate the superiority of the proposed method compared to the existing state-of-the-art conformal prediction methods.

\end{abstract}

\section{Introduction}
Uncertainty quantification has become crucial in many scientific domains where black-box machine learning models are often used~\cite{angelopoulos2021gentle}. Conformal prediction has emerged as a popular and modern technique for uncertainty quantification by providing valid predictive inference for those black-box models~\cite{shafer2008tutorial,barber2023conformal}.

Time series prediction aims to forecast future values based on a sequence of observations sequentially ordered in time~\cite{box2015time}. With recent advances in machine learning, numerous models have been proposed and adopted for various time series prediction tasks. The increased use of black-box machine learning models necessitates uncertainty quantification, particularly in high-stakes time series prediction tasks such as medical event prediction, stock prediction, and weather forecasting.

While conformal prediction can provide valid predictive inference for uncertainty quantification, applying conformal prediction to time series is challenging since time series data often violate the exchangeability assumption. Additionally, real-world time series data typically exhibit significant stochastic variations and strong temporal correlations. Many efforts have been made to develop valid and effective conformal prediction methods for time series~\cite{xu2023conformal}. Sequential Predictive Conformal Inference (SPCI), a recently proposed conformal prediction framework for time series, has shown state-of-the-art performance by using Quantile Random Forest~\cite{meinshausen2006quantile} as a conditional quantile estimator to predict the quantiles of future prediction residuals, which are used to estimate prediction interval~\cite{xu2023sequential}.

In this study, we employed Transformer decoder~\cite{vaswani2017attention,radford2018improving} as a conditional quantile estimator in the SPCI framework. Specifically, the Transformer decoder takes a sequence of past residuals and features as input to predict the quantiles of future residuals. Given that Transformer decoder-only architecture has already shown impressive performance in many sequential modeling tasks, we hypothesize that utilizing it in the SPCI framework benefits the estimation of prediction interval by learning temporal dependencies between the residuals. We empirically demonstrate the superiority of the proposed method through experiments with simulated and real data, comparing it to state-of-the-art conformal prediction methods.

\begin{figure}[t]
    \centering
    \includegraphics{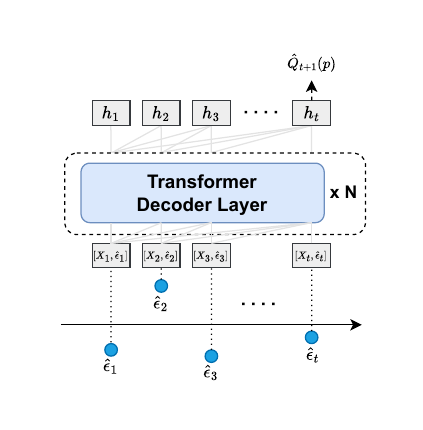}
    \vspace*{-8mm}
    \caption{Visual description of our proposed method. Prediction residuals ($\hat{\epsilon}_t$) and corresponding features ($X_t$) are used as input to a stack of Transformer decoder layers to predict the quantiles of future residual(s), which are used to estimate prediction intervals.}
    \label{fig:transformer_architecture}
\end{figure}

\section{Problem setup}

Consider a sequence of observations $\{ (X_t, Y_t) : t=1,2,\ldots\}$, where $X_t \in \mathbb{R}^d$ denotes $d$-dimensional feature and $Y_t \in \mathbb{R}$ denotes continuous scalar outcome. Assume that the first $T$ samples
$\{(X_t, Y_t)\}_{t=1}^{T}$ be the data for training and validation. We also assume that we have a point prediction method $\hat{f}$ that provides a point prediction $\hat{Y}_t$ for $Y$ as $\hat{Y}_t = \hat{f}(X_t)$.

Our goal is to sequentially construct a prediction interval $\hat{C}_{t-1} (X_t)$, starting from time $T+1$, that desirably contains the true outcome $Y_t$ with the probability at least $1-\alpha$. $\hat{C}_{t-1} (X_t)$ is constructed using the past observations and predictions up to $t-1$ time. The \textit{significance level} $\alpha$ is pre-defined and $\hat{C}_{t-1} (X_t)$ is a set includes $Y_t$ with probability at least $1-\alpha$.

We use prediction residual (i.e., prediction error) as a non-conformity score, which is defined as:
\begin{equation}
    \hat{\epsilon} = Y_t - \hat{Y}_t.
\end{equation}

Two types of coverage guarantees should be satisfied with prediction intervals: \textit{marginal coverage} and \textit{conditional coverage}.
\textit{Marginal coverage} is defined as follows:
\begin{equation}
    \mathbb{P} \left ( {Y_t} \in \hat{C}_{t-1} (X_t) \right ) \geq 1 - \alpha, \forall t.
\label{eq:marginal_coverage}
\end{equation}
\textit{Conditional coverage} is a stronger guarantee that given each $X_t$, the true observation $Y_t$ in included in $\hat{C}_{t-1} (X_t)$ with at least $1-\alpha$ probability, which can be defined as follows:
\begin{equation}
    \mathbb{P} \left ( {Y_t} \in \hat{C}_{t-1} (X_t) | X_t \right ) \geq 1 - \alpha, \forall t.
\label{eq:conditional_coverage}
\end{equation}

If $\hat{C}_{t-1} (X_t)$ satisfies eq (\ref{eq:marginal_coverage}) or eq (\ref{eq:conditional_coverage}), it is called \textit{marginally valid} or \textit{conditionally valid}, respectively. The desired aim is to construct $\hat{C}_{t-1} (X_t)$ that satisfies both marginally and conditionally valid. While infinite-width prediction intervals always satisfy both coverage guarantees, such intervals are pointless since they do not carry any information to quantify uncertainty. Therefore, we aim to minimize the width of prediction intervals while satisfying coverage.

\section{Method}

\subsection{Sequential Predictive Conformal Inference}

Sequential Predictive Conformal Inference (SPCI) is a conformal prediction framework for time series proposed by~\cite{xu2023sequential}. SPCI adopted Quantile Random Forest~\cite{meinshausen2006quantile} as a conditional quantile estimator to estimate $\hat{\epsilon}_t$ sequentially, leveraging the dependencies of the past residuals. Specifically, SPCI estimates prediction intervals as follows:
\begin{equation}
    \left[ \hat{f}(X_t) + \hat{Q}_t(\hat{\beta}) ,   \hat{f}(X_t) + \hat{Q}_t(1 - \alpha + \hat{\beta}) \right],
\end{equation}
where $\hat{Q}$ denotes a conditional quantile estimator, $\hat{Q}_t(p)$ is the estimation for the true $p$-th quantile of $\hat{\epsilon}_t$. $\hat{\beta}$ denotes the value that minimizes the prediction interval as: 
\begin{equation*}
    \hat{\beta} = \arg \min_{\beta \in [0,\alpha]} \left (\hat{Q}_t(1 - \alpha + \hat{\beta}) - \hat{Q}_t(\beta)  \right ).
\end{equation*}

\begin{figure}[tp]
    \begin{subfigure}[b]{0.45\columnwidth}
        \includegraphics[width=\columnwidth]{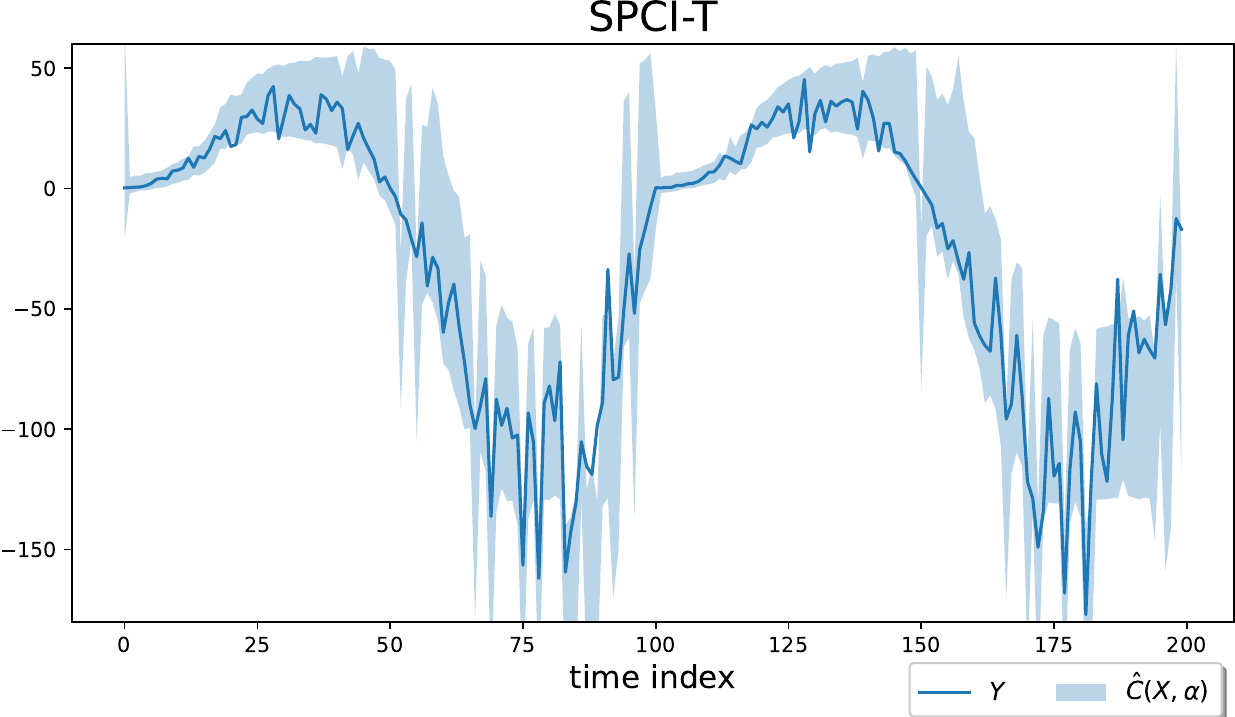}
        \caption{SPCI-T}
    \end{subfigure}
    \begin{subfigure}[b]{0.45\columnwidth}
        \includegraphics[width=\columnwidth]{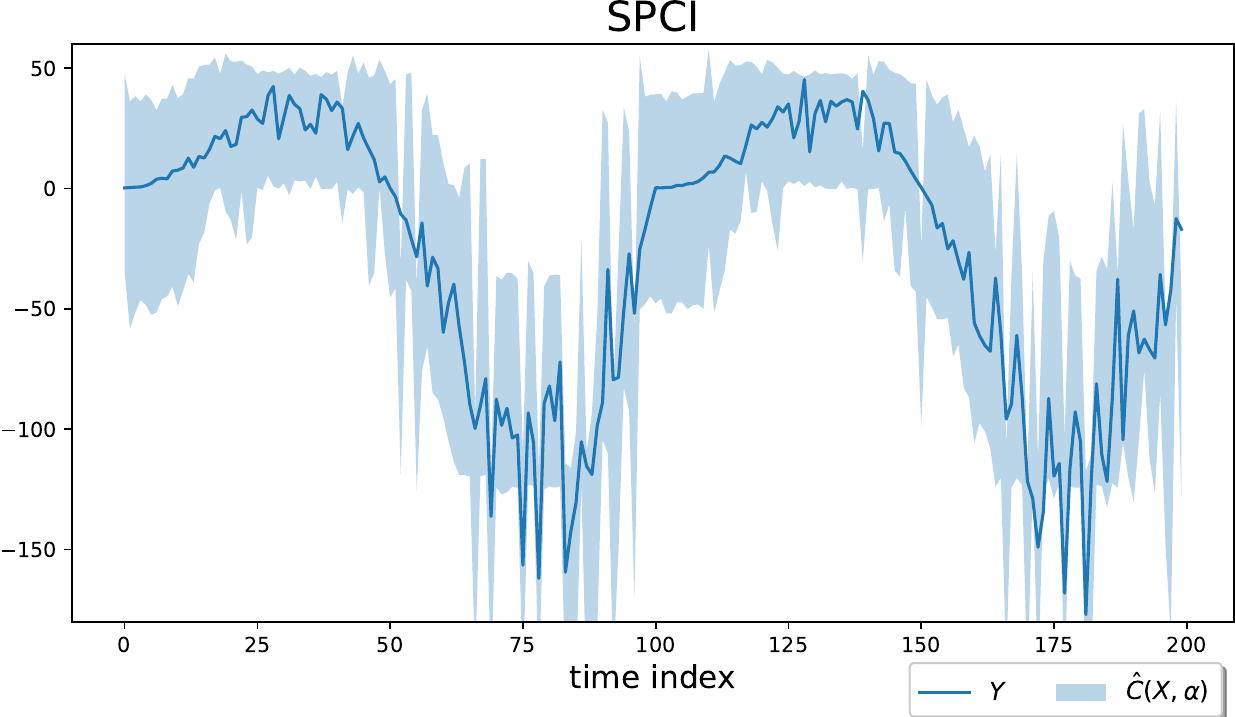}
        \caption{SPCI}
    \end{subfigure}
    \\
    \begin{subfigure}[b]{0.45\columnwidth}
        \includegraphics[width=\columnwidth]{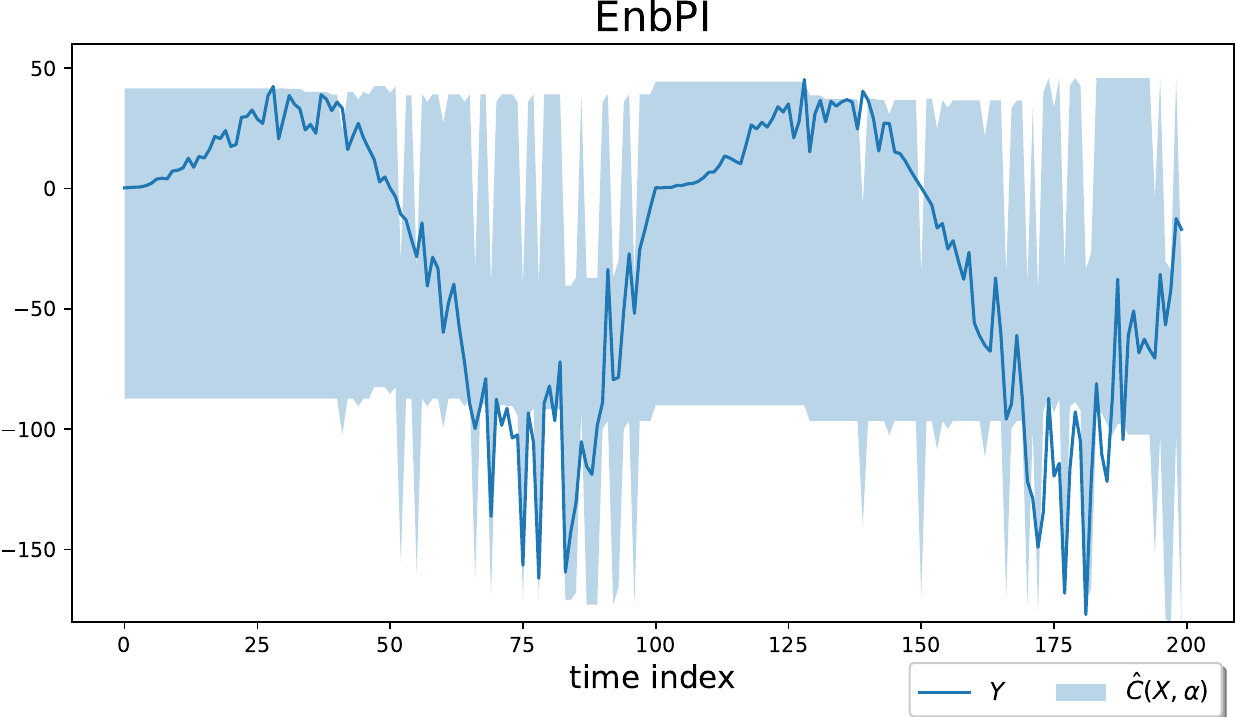}
        \caption{EnbPI}
    \end{subfigure}
    \begin{subfigure}[b]{0.45\columnwidth}
        \includegraphics[width=\columnwidth]{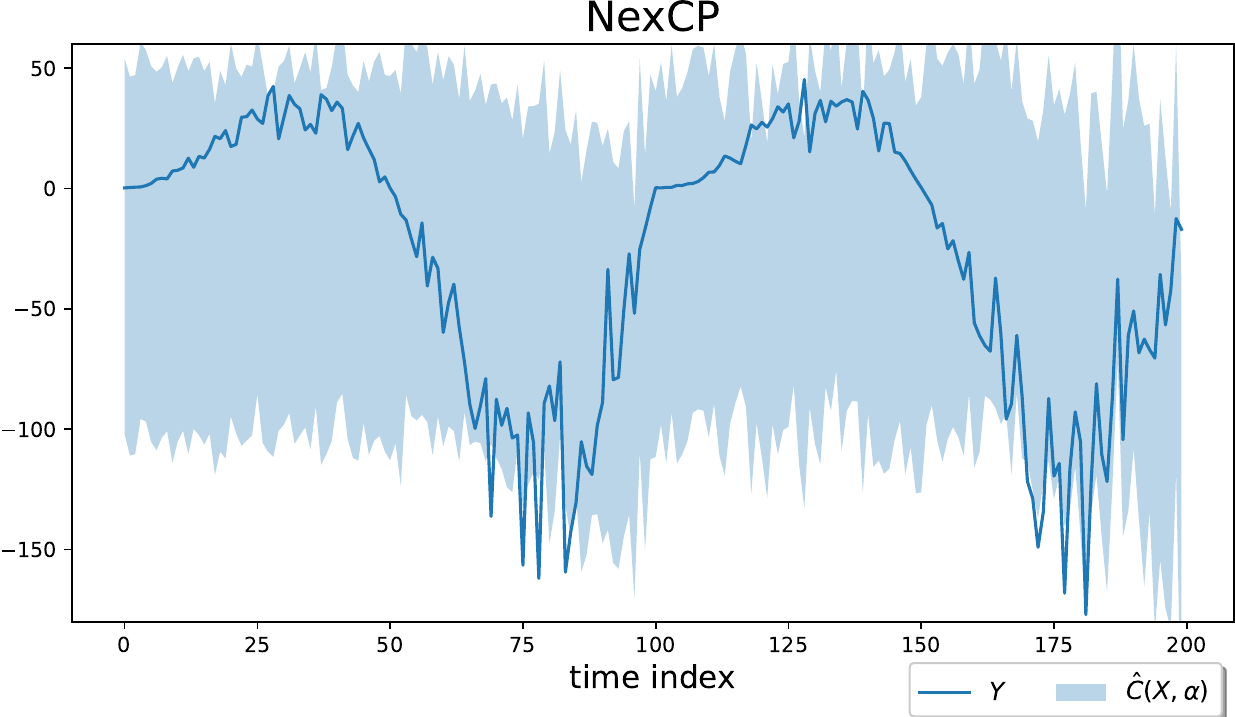}
        \caption{NexCP}
    \end{subfigure}

    \caption{Prediction intervals estimated by \texttt{SPCI-T} and baselines on non-stationary simulated dataset.}
    \label{fig:prediction_intervals}
\end{figure}

\subsection{Transformer Conformal Prediction for Time Series}

In this study, we employ Transformer~\cite{vaswani2017attention} as a conditional quantile estimator $\hat{Q}$ to estimate the true quantiles of $\hat{\epsilon}_t$ within the SPCI framework. Specifically, we use decoder-only architecture~\cite{radford2018improving} since it can generalize to variable lengths of sequences without strictly partitioning the sequences for encoding and decoding. Throughout the paper, we refer to the proposed method as Sequential Predictive Conformal Inference with Transformer (\texttt{SPCI-T}).

The past $w$ residuals and features are used as input for the model to predict the quantiles of the future unobserved residuals. Specifically, $ \{ Z_t \}_{t-(w+1)}^{t-1}$ is used to predict $\hat{Q}_t(p)$, where $Z_t := [X_t,\hat{\epsilon}_t]$. A fully connected layer without activation converts $Z_t$ to the input representation of the model dimension. Note that $Z_t$ can include other features that are useful for prediction besides $X_t$. Figure~\ref{fig:transformer_architecture} visually describes the model architecture. 

A fully connected layer without activation transforms output representation into the prediction of the quantile of $\hat{\epsilon_t}$.
Training is done by sequentially minimizing the quantile loss as follows:
\begin{equation}
\mathcal{L}(\hat{\epsilon}, \hat{\epsilon}', p) = \begin{cases}
             p (\hat{\epsilon}- \hat{\epsilon}') & \text{if } \hat{\epsilon}- \hat{\epsilon}' \geq 0, \\
             (1-p) (\hat{\epsilon}'-\hat{\epsilon})  & \text{if } \hat{\epsilon}'-\hat{\epsilon} \geq 0,
       \end{cases}
\end{equation}
where $p$ is the target quantile and $\hat{\epsilon}'$ is the predicted value of $\hat{\epsilon}$ corresponding to the target quantile.

We hypothesize that using the Transformer decoder as a conditional quantile estimator can offer the following advantages:
\begin{itemize}
    \item Transformer decoder can effectively learn temporal dependencies, including long-term dependencies, across residuals.

    \item We can incorporate additional features, such as $X_t$, for conditional quantile estimation, allowing the Transformer to learn potential dependencies between these additional features and the residuals.

    \item Transformer decoder can perform multi-step predictions using known features through generative inference without needing explicit residuals as input for prediction.
\end{itemize}

\section{Experiments}

We evaluate \texttt{SPCI-T} and baselines on simulated and real data. The code for all experiments is available at \url{https://github.com/Jayaos/TCPTS}. Hyperparameters and implementation details are available in Appendix~\ref{app:data}.

\subsection{Setup}

We first obtain point predictions for all $Y_t$ and corresponding residuals $\hat{\epsilon}$ by using leave-one-out (LOO) point predictors in all experiments. We use the ensemble of 25 random forests as the LOO point predictor. We use state-of-the-art conformal prediction methods as baselines, which include SPCI~\cite{xu2023sequential}, EnbPI~\cite{xu2021conformal}, and NexCP~\cite{barber2023conformal}. We evaluate \texttt{SPCI-T} and the baselines regarding interval coverage and width. We also evaluate the multi-step prediction of \texttt{SPCI-T}. In a multi-step prediction setup, we aim to estimate the prediction intervals at $s$-step ahead, assuming we only have known features ($X_t$) for multi-step prediction.

\subsection{Simulation}

\paragraph{Dataset} We generate two simulated time series datasets. The data generating process follows $Y_t = f(X_t) + \epsilon_t$. The first dataset is a non-stationary time series, containing periodicity and autoregressive $\epsilon_t$. The second dataset is time series with heteroscedastic errors where the variance of $\epsilon_t$ depends on $X_t$. Details on simulated data are provided in Appendix~\ref{app:data}.

\begin{table}[t]
\caption{Empirical coverage and interval width of \texttt{SPCI-T} and baselines on the simulated datasets. The target coverage is set to 0.9, and the past window is set to 100. We report the average value with standard deviation calculated based on three independent trials with different random seeds.}
\label{tb:results_simulation}
\vskip 0.15in
\begin{center}
\begin{small}
\begin{tabular}{lcccc}
\toprule
 & \multicolumn{2}{c}{Non-stationary} & \multicolumn{2}{c}{Heteroskedasticity}  \\
  & Coverage & Width &  Coverage &  Width \\
\midrule
SPCI-T & $0.91_{\pm .008}$ & $\textbf{52.38}_{\pm .584}$ & $0.88_{\pm .002}$ & $\textbf{9.56}_{\pm .082}$ \\
SPCI  & $0.96_{\pm .004}$ & $76.35_{\pm .527}$ & ${0.89}_{\pm .006}$ & ${10.03}_{\pm .001}$ \\
EnbPI &${0.86}_{\pm .002}$ & $134.7_{\pm .743}$ & ${0.90}_{\pm .006}$ & ${10.72}_{\pm .008}$ \\
NexCP & ${0.91}_{\pm .002}$ & ${156.4}_{\pm 2.73}$ & ${0.92}_{\pm .004}$ & ${11.73}_{\pm .198}$\\
\bottomrule
\end{tabular}
\end{small}
\end{center}
\vskip -0.1in
\end{table}

\begin{table}[t]
\caption{Empirical coverage and interval width of multi-step prediction using \texttt{SPCI-T} on the simulated datasets. The target coverage is set to 0.9, and the past window is set to 100.}
\label{tb:multistep_simulation}
\vskip 0.15in
\begin{center}
\begin{small}
\begin{tabular}{lcccc}
\toprule
 & \multicolumn{2}{c}{Non-stationary} & \multicolumn{2}{c}{Heteroskedasticity}  \\
  & Coverage & Width &  Coverage &  Width \\
\midrule
$s=2$ & $0.88_{\pm .015}$ & $51.54_{\pm .314}$ & $0.84_{\pm .003}$  & $9.67_{\pm .134}$ \\
$s=3$ & $0.86_{\pm .008}$ & $51.46_{\pm .183}$&  $0.80_{\pm .003}$ & $9.72_{\pm .151}$ \\
$s=4$ & $0.84_{\pm .010}$  & $51.49_{\pm .256}$&  $0.80_{\pm .010}$ & $9.76_{\pm .140}$ \\
\bottomrule
\end{tabular}
\end{small}
\end{center}
\vskip -0.1in
\end{table}

\paragraph{Results}
Table~\ref{tb:results_simulation} shows empirical coverage and interval width of \texttt{SPCI-T} and baselines on the two simulated datasets. We observe that \texttt{SPCI-T} achieves the narrowest interval width without losing coverage. Figure~\ref{fig:prediction_intervals} displays the prediction interval estimated by \texttt{SPCI-T} and baselines on the non-stationary dataset, confirming that \texttt{SPCI-T} obtains significantly narrower interval width compared to the baselines. Table~\ref{tb:multistep_simulation} presents multi-step prediction results of \texttt{SPCI-T}. While \texttt{SPCI-T} maintains its prediction interval width, it loses coverage for multi-step prediction with larger $s$.

\begin{table*}[t]
\caption{Empirical coverage and interval width of the proposed method and baselines. The target coverage is 0.9, and the past window is set to 50 for all experiments. We report the average value with standard deviation calculated based on three independent trials with different random seeds.}
\label{tb:results_real_w50}
\vskip 0.15in
\begin{center}
\begin{small}
\begin{tabular}{lcccccc}
\toprule
 & \multicolumn{2}{c}{Wind} & \multicolumn{2}{c}{Electricity} & 
\multicolumn{2}{c}{Solar} \\
  & Coverage & Width &  Coverage &  Width & Coverage & Width \\
\midrule
SPCI-T & $0.93_{\pm .006}$ & $\textbf{2.08}_{\pm .072}$ & $0.92_{\pm .009}$ & $\textbf{0.18}_{\pm .005}$ & $0.93_{\pm .005}$ & $\textbf{50.70}_{\pm 3.84}$ \\
SPCI  & $0.96_{\pm .016}$ & $2.41_{\pm .016}$ & $0.92_{\pm .004}$ & $0.22_{\pm .001}$ & $0.91_{\pm .006}$ & $88.76_{\pm .245}$ \\
EnbPI & $0.48_{\pm .006}$ & $4.10_{\pm .009}$ & $0.79_{\pm .002}$ & $0.22_{\pm .001}$ & 
 $0.88_{\pm .002}$ & $86.91_{\pm .363}$\\
NexCP & $0.92_{\pm .016}$ & $6.27_{\pm .145}$ & $0.89_{\pm .001}$ & $0.46_{\pm .001}$ & $0.86_{\pm .002}$ & $114.98_{\pm .201}$ \\
\bottomrule
\end{tabular}
\end{small}
\end{center}
\vskip -0.1in
\end{table*}

\begin{table*}[t]
\caption{Empirical coverage and interval width of the proposed method and baselines. The target coverage is 0.9, and the past window is set to 100 for all experiments. Note that the performance of NexCP is identical to the performance in Table~\ref{tb:results_real_w50} since it does not use the past window.}
\label{tb:results_real_w100}
\vskip 0.15in
\begin{center}
\begin{small}
\begin{tabular}{lcccccc}
\toprule
 & \multicolumn{2}{c}{Wind} & \multicolumn{2}{c}{Electricity} & 
\multicolumn{2}{c}{Solar} \\
  & Coverage & Width &  Coverage &  Width & Coverage & Width \\
\midrule
SPCI-T & $0.91_{\pm .011}$ & $\textbf{1.96}_{\pm .094}$ & $0.92_{\pm .013}$ & $\textbf{0.17}_{\pm .014}$ & $0.90_{\pm .006}$ & $\textbf{45.35}_{\pm 1.67}$ \\
SPCI  & $0.94_{\pm .006}$ & $2.39_{\pm .030}$ & $0.93_{\pm .005}$ & $0.22_{\pm .002}$ & $0.93_{\pm .002}$ & $88.21_{\pm .434}$ \\
EnbPI & $0.74_{\pm .038}$ & $4.65_{\pm .026}$ & $0.85_{\pm .001}$ & $0.26_{\pm .001}$ & $0.88_{\pm .002}$ & $86.64_{\pm .138}$ \\
NexCP & $0.92_{\pm .016}$ & $6.27_{\pm .145}$ & $0.89_{\pm .001}$ & $0.46_{\pm .001}$ & $0.86_{\pm .002}$ & $114.98_{\pm .201}$ \\
\bottomrule
\end{tabular}
\end{small}
\end{center}
\vskip -0.1in
\end{table*}

\subsection{Real Data Experiments}

\paragraph{Dataset}

We use three time-series datasets from the real world: solar, wind, and electricity. The solar dataset \cite{zhang2021solar} contains solar radiation information in Atalanta downtown measured in terms of diffuse horizontal irradiance, provided by the United States National Solar Radiation Database. The wind dataset consists of wind speed records measured by the Midcontinent Independent System Operator every 15 minutes over a week period in September 2020~\cite{zhu2021multi}. The electricity dataset contains electricity usage and
pricing in the states of New South Wales and Victoria in Australia, observed between 1996 and 1999~\cite{harries1999splice}. All three datasets were widely adopted benchmark datasets in conformal prediction literature. Details of the three datasets are provided in Appendix~\ref{app:data}.

\paragraph{Results}

Table~\ref{tb:results_real_w50} and Table~\ref{tb:results_real_w100} shows empirical coverage and interval width of \texttt{SPCI-T} and baselines on three real datasets with $w=50$ and $w=100$, respectively. We observe that \texttt{SPCI-T} consistently outperforms all other baselines. We also observe that the interval width of \texttt{SPCI-T} was narrow with a longer window, which shows that \texttt{SPCI-T} utilizes the long history of residuals by learning dependencies. Table~\ref{tb:multistep_realdata} shows multi-step prediction results of \texttt{SPCI-T} on electricity and solar datasets. We exclude the results on the wind dataset since the coverage deteriorated significantly with $s > 2$. Similarly, in simulation, we observed that \texttt{SPCI-T} lost the coverage with the increasing $s$.

We add additional time features to each $X_t$ in the solar dataset to see how these additional features can influence the performance of \texttt{SPCI-T}. Table~\ref{tb:solar_add_features} presents empirical coverage and interval width of \texttt{SPCI-T} and baselines on a solar dataset with additional features. \texttt{SPCI-T} again consistently outperforms baselines and shows significantly improved performance compared to the experiment on the solar dataset without the additional features. These results empirically demonstrate the advantage of \texttt{SPCI-T} in utilizing features with the residuals for conditional quantile estimation.

\begin{table}[t]
\caption{Empirical coverage and interval width of multi-step prediction using SPCI-T on real datasets. The target coverage is 0.9, and the past window is set to 100.}
\label{tb:multistep_realdata}
\vskip 0.15in
\begin{center}
\begin{small}
\begin{tabular}{lcccc}
\toprule
& \multicolumn{2}{c}{Electricity} & 
\multicolumn{2}{c}{Solar} \\
  & Coverage & Width &  Coverage &  Width \\
\midrule
$s=2$ & $0.87_{\pm .004}$ & $0.18_{\pm .007}$ & $0.88_{\pm .005}$ & $42.70_{\pm 1.20}$\\
$s=3$ & $0.82_{\pm .004}$ & $0.18_{\pm .007}$ & $0.86_{\pm .010}$& $44.56_{\pm 1.82}$\\
$s=4$ & $0.78_{\pm .007}$ & $0.18_{\pm .007}$ & $0.84_{\pm .020}$ & $46.83_{\pm 2.08}$\\
\bottomrule
\end{tabular}
\end{small}
\end{center}
\vskip -0.1in
\end{table}

\begin{table}[t]
\caption{Empirical coverage and interval width of \texttt{SPCI-T} and baselines on a solar dataset with additional features. The target coverage is 0.9, and the past window is set to 50 or 100. Note that NexCP showed the identical performance regardless of $w$ since NexCP does not use past windows.}
\label{tb:solar_add_features}
\vskip 0.15in
\begin{center}
\begin{small}
\begin{tabular}{lcccc}
\toprule
&
\multicolumn{4}{c}{Solar with additional features} \\
& \multicolumn{2}{c}{$w=50$} & \multicolumn{2}{c}{$w=100$} \\
 &  Coverage &  Width & Coverage & Width \\
\midrule
SPCI-T & $0.93_{\pm .013}$  & $\textbf{33.51}_{\pm 1.99}$  & $0.91_{\pm .006}$ & $\textbf{28.14}_{\pm 1.60}$ \\
SPCI & $0.92_{\pm .002}$  & $89.63_{\pm .651}$ &$0.92_{\pm .005}$  &  $87.28_{\pm .782}$  \\
EnbPI & $0.88_{\pm .005}$ & $86.94_{\pm .181}$ & $0.89_{\pm .000}$ & $85.73_{\pm .459}$ \\
NexCP & $0.90_{\pm .002}$ & $100.92_{\pm 6.64}$ & $0.90_{\pm .002}$ & $100.92_{\pm 6.64}$ \\
\bottomrule
\end{tabular}
\end{small}
\end{center}
\vskip -0.1in
\end{table}

\section{Conclusion}

In this study, we propose \texttt{SPCI-T}, which incorporates a Transformer decoder into the recently developed SPCI framework. \texttt{SPCI-T} uses a Transformer decoder to learn temporal dependencies between the residuals and features to predict the conditional quantile of future residuals, which are then used to estimate the prediction interval. Our simulated and real data experiments empirically demonstrate the superiority of \texttt{SPCI-T}. Future directions include tailoring the model architecture for the time series conformal prediction task and conducting more comprehensive evaluations with state-of-the-art methods.


\section*{Acknowledgments}

This work is partially supported by an NSF CAREER CCF-1650913, NSF DMS-2134037, CMMI-2015787, CMMI-2112533, DMS-1938106, DMS-1830210, and the Coca-Cola Foundation.

\vspace{-0.1in}

\bibliography{refs}

\appendix

\section{Additional Details of Experiments}
\label{app:data}

\subsection{Details of Simulated Data}

\paragraph{Non-stationary time series}

We set $T=2000$, which means we have $\{(X_t, Y_t)\}_{t=1}^{2000}$. We first generate each dimension of $X_t \in \mathbb{R}^{10}$ from $\text{Unif}(0,e^{0.01\text{mod}(t,100)})$ for all $t$. Then, $f(X_t)$ is obtained as follows:
\begin{equation*}
    f(X_t) = g(X_t) h(X_t),
\end{equation*}
where
\begin{equation*}
    g(t) = \log (t') \sin (2 \pi t' / 100),\quad t'= \text{mod}(t,100),
\end{equation*}
and
\begin{equation*}
    h(X_t) = (|\beta^{\top}X_t|+(\beta^{\top}X_t)^2 + |\beta^{\top}X_t|^3)^{1/4}.
\end{equation*}

We sample $\epsilon_t$ from AR(1) as $\epsilon_t = \rho \epsilon_{t-1} + e_t$, where $\rho=0.6$ and $e_t$ is \textit{i.i.d.} generated from normal distribution with zero mean and unit variance. $\beta \in \mathbb{R}^{10}$ is a sparse vector having only 20\% non-zero elements where non-zero elements are generated from $\text{Unif}(0,1)$.

\paragraph{Time series with heteroskedastic error} We set $T=2000$, then generate $X_t \in \mathbb{R}^{10}$ similarly to non-stationary time series for all $t$. Then, $f(X_t)$ is obtained as follows:
\begin{equation*}
    f(X_t) = (|\beta^{\top}X_t|+(\beta^{\top}X_t)^2 + |\beta^{\top}X_t|^3)^{1/4},
\end{equation*}
where $\beta$ is generated in a similar way as in the non-stationary time series. We also sample $\epsilon_t$ from AR(1) similar to non-stationary time series, but with the variance of $\epsilon_t$ dependent on $X_t$ as follows:
\begin{equation*}
    \text{Var}(\epsilon_t) = \sigma(X_t)^2,
\end{equation*}
\begin{equation*}
    \sigma(X_t) = \mathbf{1}^{\top} X_t.
\end{equation*}

\subsection{Details of Real Data}

\paragraph{Solar data}

The solar dataset contains solar radiation information recorded in every 30 minutes in 2018 in Atalanta downtown~\cite{zhang2021solar} provided
by the United States National Solar Radiation Database~\cite{sengupta2018national}. We used seven covariates: Direct Normal Irradiance (DNI), dew point, surface albedo, wind speed, relative humidity, temperature, and pressure. The outcome variable is Diffuse Horizontal Irradiance (DHI), which reflects radiation levels. For additional time features, we converted the 24-hour period into 24 hourly one-hot encoded features, adding 24 additional features.

\paragraph{Wind data}

The wind dataset contains wind speed data (measured in m/s) from wind farms operated by the Midcontinent Independent System Operator (MISO) in the United States~\cite{zhu2021multi}. This 10-dimensional dataset recorded wind speed every 15 minutes over a one-week period in September 2020.

\paragraph{Electricity data} 
The electricity dataset recorded electricity usage and pricing in the states of New South Wales and Victoria in Australia, every 30 minutes over a 2.5 year period between 1996 and 1999~\cite{harries1999splice}. We used four covariates: \textit{nswprice} and \textit{vicprice}, the price of electricity in each of the two states; and \textit{nswdemand} and \textit{vicdemand}, the usage demand in each of the two states. The outcome variable is \textit{transfer}, which is the quantity of electricity transferred between the two states.

\subsection{Hyperparameters and Implementation Details}

Table~\ref{tb:hyperparams} shows hyperparameters chosen for \texttt{SPCI-T}. We conducted a grid search using training and validation set to find the optimal hyperparameters. The model dimension, number of heads, and number of layers were chosen when performance plateaued. Since \texttt{SPCI-T} requires validation set to select the best model during training, we split the datasets into training, validation, and test set with 8:1:1 ratio for \texttt{SPCI-T}. For all other baselines, the datasets were split into training and test set with a 9:1 ratio. For a fair comparison in terms of data usage, we additionally trained \texttt{SPCI-T} on the validation set after it was initially trained on the training set, which was conducted for 10\% of the number of epochs used for the initial training.

\begin{table}[t]
\caption{Hyperparameters chosen for \texttt{SPCI-T} in experiments using simulated and real data.}
\label{tb:hyperparams}
\vskip 0.15in
\begin{center}
\begin{small}
\begin{tabular}{lcccccc}
\toprule
& \multicolumn{2}{c}{Simulation} & \multicolumn{4}{c}{Real data} \\
 &  Non-stationary &  Heteroskedastic & Wind & Electricity & Solar & Solar w/ add. features \\
\midrule
batch size &  4 & 4 & 4 & 4 & 4 & 4\\
learning rate &  0.0001 & 0.0001 & 0.0001 & 0.0001 &  0.0005 & 0.0005 \\
model dimension & 16 & 16 & 32 & 16 & 16 & 32 \\
number of heads & 4 & 4 & 4 & 4 & 4 & 4\\
number of layers & 4 & 4 & 4 & 4 & 4 & 4\\
dropout & 0.2 & 0.2 & 0.1 & 0.2 & 0.2 & 0.2 \\
additional training & \textcolor{red}{N} & \textcolor{red}{N} & \textcolor{blue}{Y} & \textcolor{blue}{Y} & \textcolor{blue}{Y} & \textcolor{blue}{Y}\\

\bottomrule
\end{tabular}
\end{small}
\end{center}
\vskip -0.1in
\end{table}

\end{document}